\documentclass[lettersize,journal]{IEEEtran}
\usepackage{tikz,xcolor,hyperref}
\definecolor{lime}{HTML}{A6CE39}
\DeclareRobustCommand{\orcidicon}{%
    \begin{tikzpicture}
    \draw[lime, fill=lime] (0,0) 
    circle [radius=0.16] 
    node[white] {{\fontfamily{qag}\selectfont \tiny ID}};    \draw[white, fill=white] (-0.0625,0.095) 
    circle [radius=0.007];    \end{tikzpicture}
    \hspace{-2mm}}
\foreach \x in {A, ..., Z}{%
    \expandafter\xdef\csname orcid\x\endcsname{\noexpand\href{https://orcid.org/\csname orcidauthor\x\endcsname}{\noexpand\orcidicon}}
    }

\usepackage{amsmath,amsfonts}
\usepackage{algorithm}
\usepackage{algpseudocode}
\usepackage{array}
\usepackage{pifont}
\usepackage{textcomp}
\usepackage{stfloats}
\usepackage{url}
\usepackage{verbatim}
\usepackage{graphicx}
\usepackage{cite}
\usepackage{amsmath}
\usepackage{amssymb}
\usepackage{mathtools}
\usepackage{amsthm}
\usepackage{hyperref}
\usepackage{url}
\usepackage[utf8]{inputenc} 
\usepackage[T1]{fontenc}    
\usepackage{hyperref}       
\usepackage{url}            
\usepackage{booktabs}       
\usepackage{amsfonts}       
\usepackage{nicefrac}       
\usepackage{microtype}      
\usepackage{xcolor}         
\usepackage{graphicx}
\usepackage{wrapfig}
\usepackage{multirow}
\usepackage{soul}
\usepackage{makecell}
\usepackage{graphicx}
\usepackage{subcaption}
\usepackage{comment}
\usepackage{threeparttable}
\usepackage{float}
\usepackage{bm}
\usepackage[english]{babel}
\def\BibTeX{{\rm B\kern-.05em{\sc i\kern-.025em b}\kern-.08em
    T\kern-.1667em\lower.7ex\hbox{E}\kern-.125emX}}
\usepackage{balance}

\hypersetup{
	colorlinks=true,
	linkcolor=red,     
	citecolor=blue,     
	filecolor=blue,     
	urlcolor=blue       
}

\begin{document}
\title{\includegraphics[width=0.05\textwidth]{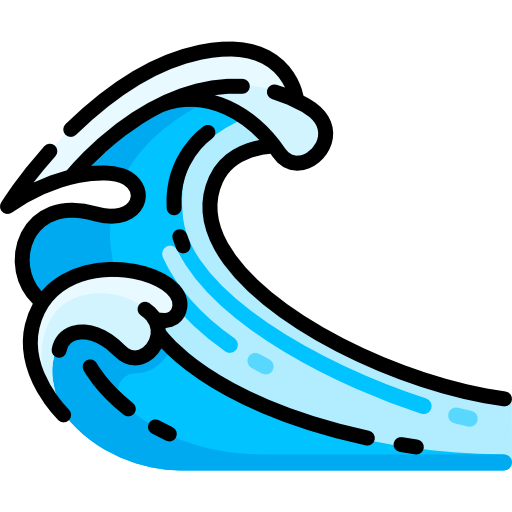}UWBench: A Comprehensive Vision-Language Benchmark for Underwater Understanding}
\author{
	Da~Zhang\orcidA{},~\IEEEmembership{Student Member,~IEEE,}
Chenggang Rong\orcidR{},
Bingyu~Li\orcidB{}, 
Feiyu Wang\orcidW{},
Zhiyuan~Zhao\orcidC{},
Junyu~Gao\orcidD{},~\IEEEmembership{Member,~IEEE,}
~and~
Xuelong~Li\orcidF{}$^{\dagger}$,~\IEEEmembership{Fellow,~IEEE}

\thanks{Da Zhang, Chenggang Rong, and Junyu Gao are with the Institute of Artificial Intelligence (TeleAI), China Telecom, China and also
with the School of Artificial Intelligence, OPtics and ElectroNics (iOPEN), Northwestern Polytechnical University, Xi'an 710072, China. 
(E-mail: dazhang@mail.nwpu.edu.cn; rongcg5620@mail.nwpu.edu.cn; gjy3035@gmail.com).}

\thanks{
	Bingyu Li, Feiyu Wang, Zhiyuan Zhao, and 
	Xuelong Li is with the Institute of Artificial Intelligence (TeleAI), China Telecom, China. 
	(E-mail: 
	libingyu0205@mail.ustc.edu.cn; wangfy25@m.fudan.edu.cn; tuzixini@gmail.com;
xuelong\_li@ieee.org).}

}


\maketitle

\begin{abstract}
Large vision-language models (VLMs) have achieved remarkable success in natural scene understanding, yet their application to underwater environments remains largely unexplored. Underwater imagery presents unique challenges including severe light attenuation, color distortion, and suspended particle scattering, while requiring specialized knowledge of marine ecosystems and organism taxonomy. To bridge this gap, we introduce UWBench, a comprehensive benchmark specifically designed for underwater vision-language understanding. 
UWBench comprises 15,003 high-resolution underwater images captured across diverse aquatic environments, encompassing oceans, coral reefs, and deep-sea habitats. 
Each image is enriched with human-verified annotations including 15,281 object referring expressions that precisely describe marine organisms and underwater structures, and 124,983 question-answer pairs covering diverse reasoning capabilities from object recognition to ecological relationship understanding. 
The dataset captures rich variations in visibility, lighting conditions, and water turbidity, providing a realistic testbed for model evaluation. 
Based on UWBench, we establish three comprehensive benchmarks: detailed image captioning for generating ecologically informed scene descriptions, visual grounding for precise localization of marine organisms, and visual question answering for multimodal reasoning about underwater environments. Extensive experiments on state-of-the-art VLMs demonstrate that underwater understanding remains challenging, with substantial room for improvement. Our benchmark provides essential resources for advancing vision-language research in underwater contexts and supporting applications in marine science, ecological monitoring, and autonomous underwater exploration. Our code and benchmark are available at \href{https://github.com/zhangda1018/UWBench}{UWBench}.

\end{abstract}


\section{Introduction}
\label{sec:intro}

\IEEEPARstart{R}{ecent} years have witnessed remarkable advances in large vision-language models (VLMs) \cite{xu2024lvlm, zhang2024vision, li2025benchmark}, which have demonstrated unprecedented capabilities in understanding and reasoning about visual content through natural language \cite{gong2025figstep, zhu2025ibd}. 
State-of-the-art VLMs such as GPT-5, GLM-4.5, and InternVL have achieved impressive performance across a wide range of tasks, including detailed image captioning \cite{li2025underwater, sarto2025image, hua2025finecaption}, complex visual question answering \cite{huang2025frames, manas2024improving}, and precise visual grounding \cite{xiao2024towards, liu2025survey}. 
These models have enabled significant progress by effectively bridging the gap between vision and language, and have been successfully deployed in various real-world applications, ranging from autonomous navigation \cite{nahavandi2025comprehensive, lee2024learning} to content moderation \cite{kolla2024llm, palla2025policy} and assistive technologies \cite{yuan2024towards, padmanabha2024voicepilot}. 
Much of this progress can be attributed to the availability of large-scale, high-quality datasets captured in terrestrial environments \cite{li2025openhumanvid, albalak2025big, wang2025koala}, which have provided VLMs with rich and diverse data for training and evaluation in natural scenes.
However, a critical question remains largely unanswered: \textbf{can current VLMs effectively understand and interpret imagery from challenging underwater environments, where visual conditions differ fundamentally from terrestrial scenarios?}

\begin{figure}[t]
	\centering
	\includegraphics[width=1\linewidth]{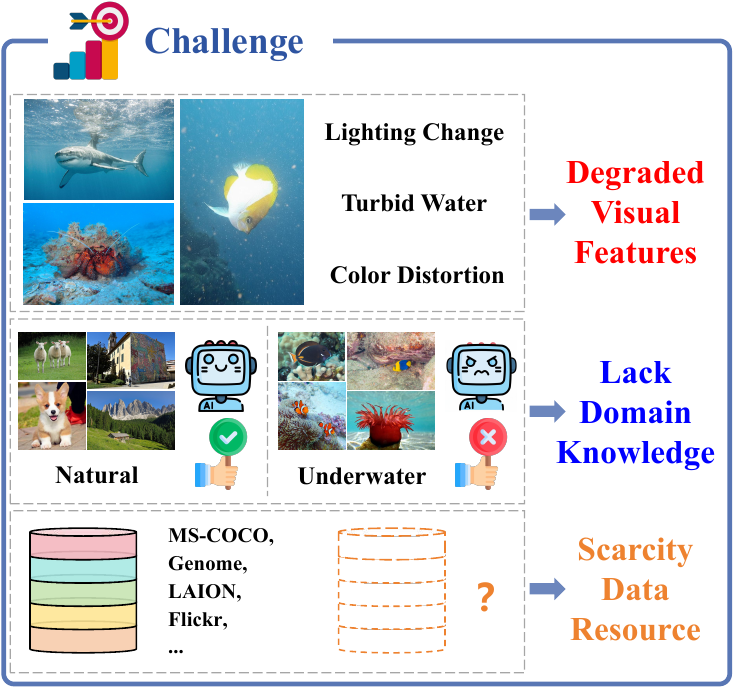} 
	\caption{Challenges for VLMs for understanding underwater scenes.
	}
	\label{fig1}

\end{figure}

This question carries substantial importance given the vital role of underwater observation in marine science, ecological conservation, and resource management \cite{pei2024research, chen2024global, spalding2025state}. 
As illustrated in Figure \ref{fig1}, applying VLMs to underwater imagery introduces three fundamental challenges that distinguish this domain from conventional vision-language tasks:

\begin{itemize}

    \item \textbf{Degraded visual features}. Underwater scenes exhibit variable illumination with rapid light attenuation, wavelength-dependent color distortion, and fluctuating turbidity from suspended particles, resulting in reduced contrast and limited visibility that confound terrestrial-trained approaches \cite{zhou2025spatial, wang2023domain}. 
    
    \item \textbf{Lack of scientific domain knowledge}. Accurate underwater interpretation requires specialized knowledge of marine taxonomy, organism morphology, behavioral patterns, and ecological relationships, representing a substantial gap for systems trained on common terrestrial scenarios \cite{wu2024self}. 
    
    \item \textbf{Scarcity of data resources}. Existing vision-language datasets focus predominantly on terrestrial scenarios, while underwater-specific datasets lack the multimodal annotations necessary for comprehensive understanding \cite{fuad2025aqua20}.

\end{itemize}

\begin{figure*}[t]
	\centering
	\includegraphics[width=1\linewidth]{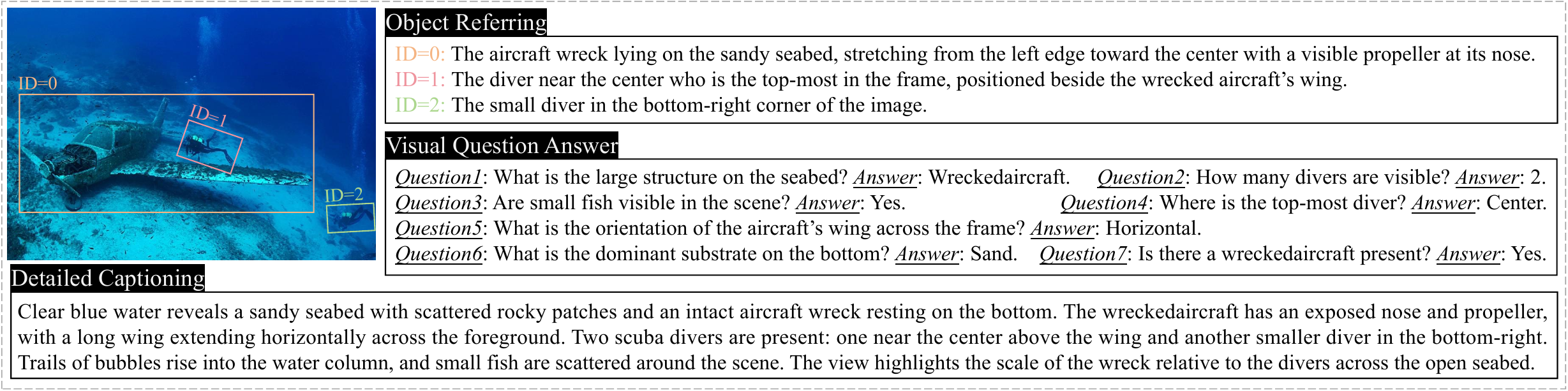} 
	\caption{Examples of an image and corresponding annotations in UWBench dataset. Our annotations include object referring, visual question answering, and detailed captions.
	}
	\label{fig2}
	\vspace{-0.2cm}
\end{figure*}

To address these challenges and advance vision-language understanding in underwater contexts, we introduce UWBench, the first large-scale benchmark dataset specifically designed for comprehensive underwater image understanding. UWBench comprises 15,003 high-resolution images spanning diverse underwater environments including shallow reefs, mid-depth zones, and deep-sea habitats, with rich variations in water clarity, illumination conditions, and marine biodiversity that reflect realistic observation scenarios. Each image is enriched with meticulously crafted annotations verified by marine biology experts: 15,003 detailed captions providing ecologically informed scene descriptions, 15,281 object referring expressions for precise organism localization, and 124,983 question-answer pairs assessing diverse reasoning capabilities from basic recognition to complex ecological understanding (Figure \ref{fig2}). Through a collaborative approach combining automated GPT-5 generation with rigorous human verification, we ensure scientific accuracy in taxonomic classifications, behavioral descriptions, and environmental context. Based on these comprehensive annotations, UWBench establishes three interconnected evaluation benchmarks for detailed image captioning, visual grounding, and visual question answering, enabling systematic assessment of vision-language models across diverse underwater understanding capabilities and supporting development of specialized systems for marine research and ecological monitoring applications.

 We conduct comprehensive experiments on UWBench to evaluate the performance of existing VLMs under three evaluation settings. 
 Specifically, we assess multiple state-of-the-art models including closed-source systems such as GPT-4o \cite{achiam2023gpt}, GPT-5, and Gemini, as well as open-source models including InternVL3.5 series \cite{wang2025internvl3}, Qwen2.5-VL series \cite{hui2024qwen2}, Qwen3 series \cite{yang2025qwen3}, and GLM-4 series \cite{glm2024chatglm}. 
 Experimental results demonstrate that underwater image understanding remains highly challenging even for state-of-the-art models. 
 Compared to their performance on terrestrial benchmarks, all models exhibit substantial performance degradation on UWBench, highlighting the domain gap between terrestrial and underwater vision-language understanding. 
 Through comprehensive analysis including task-specific evaluation, attribute-based assessment, and qualitative error analysis, we identify key factors contributing to performance decline: difficulty in recognizing degraded visual features under poor visibility conditions, insufficient domain knowledge for accurate taxonomic identification and ecological reasoning, and challenges in fine-grained spatial reasoning required for precise object localization in complex underwater scenes. 
 These findings underscore the necessity of specialized benchmarks like UWBench and illuminate potential avenues for future improvement in underwater vision-language understanding. The key contributions of our work are summarized as follows:
 
 \begin{itemize}
 \item We construct UWBench, the first large-scale benchmark dataset specifically designed for underwater vision-language understanding, comprising 15,003 high-resolution images with comprehensive human-verified annotations including detailed captions, object referring expressions, and question-answer pairs.

 \item Three comprehensive evaluation benchmarks are established for detailed image captioning, visual grounding, and visual question answering, providing standardized protocols for systematic assessment of vision-language models in underwater contexts.

 
 \item We provide detailed insights into factors limiting current model performance and identify promising directions for future research, contributing to the development of specialized vision-language systems capable of supporting marine research and ecological monitoring applications.

%
%


\end{itemize}

\section{Related Works}
\label{related_works}

\subsection{Vision-Language Models and Benchmarks}

The rapid advancement of large vision-language models has revolutionized multimodal understanding across diverse applications \cite{yin2024survey, awais2025foundation}. 
Foundational works such as CLIP \cite{radford2021learning} demonstrate the effectiveness of contrastive learning for aligning visual and textual representations at scale \cite{yang2024depth, simeoni2025dinov3}. 
Building upon these foundations, recent generative VLMs including LLaVA \cite{li2024llava}, InternVL \cite{chen2024internvl}, Qwen-VL \cite{wang2024qwen2}, and GLM \cite{glm2024chatglm} have achieved remarkable performance by integrating powerful vision encoders with large language models, enabling sophisticated reasoning about visual content through natural language. These models excel at tasks requiring joint understanding of vision and language, including image captioning \cite{zheng2024dreamlip}, visual question answering \cite{huang2025frames}, and visual grounding \cite{tang2023context}.

The development of comprehensive benchmarks has been instrumental in advancing VLM capabilities \cite{xu2024lvlm, yang2025wildvideo, hong2025lvos}. 
Datasets such as COCO \cite{lin2014microsoft} provide large-scale annotations for object detection and image captioning, while VQAv2 \cite{goyal2017making} and OK-VQA \cite{marino2019ok} introduce challenging visual question answering scenarios requiring reasoning and commonsense knowledge. 
RefCOCO \cite{chen2025revisiting} and its variants establish benchmarks for referring expression comprehension and visual grounding tasks. 
More recently, specialized benchmarks like MMBench \cite{liu2024mmbench} and SEED-Bench \cite{li2023seed} provide holistic evaluation frameworks assessing multiple dimensions of multimodal understanding. 
However, these benchmarks predominantly focus on everyday terrestrial scenarios, leaving significant gaps in specialized domains. UWBench addresses this limitation by providing the first comprehensive vision-language benchmark specifically designed for underwater environments, enabling systematic evaluation of VLM capabilities in challenging aquatic contexts.

\subsection{Underwater Image Datasets and Benchmarks}

Underwater vision research has progressed through the development of specialized datasets targeting various perception tasks \cite{li2019underwater, islam2020fast, raveendran2021underwater}. Early efforts focused on low-level image enhancement \cite{wang2024metalantis} and restoration \cite{shen2024u2pnet}. 
The UIEBD dataset \cite{li2019underwater} provides paired underwater and reference images for enhancement algorithm evaluation, while EUVP \cite{islam2020fast} offers a large-scale collection for underwater image restoration and color correction. More recent datasets like LSUI \cite{peng2023u} and RUIE \cite{liu2020real} extend this work with diverse underwater scenes captured under varying environmental conditions.

For object-level understanding, several datasets have been proposed to advance underwater object detection and segmentation. The URPC dataset \cite{liu2021dataset} introduces annotations for marine organisms in challenging underwater conditions, while Brackish \cite{pedersen2023brackishmot} includes diverse aquatic species for detection and tracking. 
SUIM \cite{islam2020semantic} provides semantic segmentation annotations for underwater imagery, and more recently, UIIS \cite{lian2023watermask} offers instance-level segmentation masks for marine objects. 
The USOD10K dataset \cite{hong2023usod10k} marks significant progress by providing 10,255 images with pixel-wise annotations for underwater salient object detection, covering 70 object categories across diverse underwater environments. 
Building upon this foundation, USIS16K \cite{hong2025usis16k} further expands the scale to 16,151 images with 158 categories, offering comprehensive instance segmentation annotations.

For temporal understanding, tracking datasets such as UTB180 \cite{alawode2022utb180} and VMAT \cite{cai2023semi} provide annotated video sequences for single object tracking in underwater scenarios. More recently, WebUOT-1M \cite{zhang2024webuot} introduces a million-scale benchmark with 1,500 video sequences spanning 408 categories, significantly advancing underwater object tracking research. 
These datasets have substantially contributed to advancing underwater computer vision. However, they primarily focus on traditional vision tasks without incorporating vision-language annotations. In contrast, UWBench bridges this critical gap by providing comprehensive multimodal annotations including detailed captions, referring expressions, and question-answer pairs, thereby enabling the development and evaluation of vision-language models specifically tailored for underwater understanding. Table \ref{table1} provides a comprehensive comparison of representative underwater datasets across different task categories.

\begin{table*}[htbp]
	\centering
	\caption{Comprehensive comparison of underwater datasets and benchmarks. UWBench is the first image-based benchmark providing integrated vision-language annotations including detailed captions, referring expressions, and visual question answering, alongside traditional detection and segmentation tasks.}
	\resizebox{\textwidth}{!}{
		\begin{tabular}{lccccccccccccccc}
			\toprule
			Dataset & Year & Type & Images & Videos & Categories & Annotation Quality & Enhancement & Detection & Segmentation & Tracking & Caption & Referring & VQA \\
			\midrule
			UIEBD \cite{li2019underwater} & 2019 & Image & 950 & - & - & Manual & \ding{51} & - & - & - & - & - & - \\
			EUVP \cite{islam2020fast} & 2020 & Image & 12K & - & - & Manual & \ding{51} & - & - & - & - & - & - \\
			LSUI \cite{peng2023u} & 2023 & Image & 5K & - & 10 & Manual & \ding{51} & - & - & - & - & - & - \\
			RUIE \cite{liu2020real} & 2021 & Image & 4K & - & - & Semi-auto & \ding{51} & - & - & - & - & - & - \\
			URPC \cite{liu2021dataset} & 2020 & Image & 6K & - & 4 & Manual & - & \ding{51} & - & - & - & - & - \\
			Brackish \cite{pedersen2023brackishmot} & 2020 & Image & 14K & 89 & 8 & Manual & - & \ding{51} & - & \ding{51} & - & - & - \\
			SUIM \cite{islam2020semantic} & 2020 & Image & 1.6K & - & 8 & Manual & - & - & \ding{51} & - & - & - & - \\
			UIIS \cite{lian2023watermask} & 2023 & Image & 4.6K & - & 7 & Manual & - & - & \ding{51} & - & - & - & - \\
			USOD10K \cite{hong2023usod10k}& 2023 & Image & 10.3K & - & 70 & Manual & - & - & \ding{51} & - & - & - & - \\
			USIS16K \cite{hong2025usis16k}& 2025 & Image & 16.2K & - & 158 & Manual+Expert & - & \ding{51} & \ding{51} & - & - & - & - \\
			DeepFish \cite{qin2016deepfish} & 2022 & Image & 4.5K & - & 20 & Manual & - & \ding{51} & \ding{51} & - & - & - & - \\
			FishNet \cite{khan2023fishnet}& 2023 & Image & 17K & - & 17K & Manual & - & \ding{51} & - & - & - & - & - \\
			UTB180 \cite{alawode2022utb180} & 2018 & Video & - & 180 & - & Manual & - & - & - & \ding{51} & - & - & - \\
			VMAT \cite{cai2023semi} & 2023 & Video & 57K & 33 & 17 & Manual & - & - & - & \ding{51} & - & - & - \\
			WebUOT-1M \cite{zhang2024webuot}& 2024 & Video & 1M & 1.5K & 408 & Semi-auto & - & - & - & \ding{51} & - & - & - \\
			DRUVA \cite{varghese2023self}& 2023 & Video & 6K & 20 & 20 & Manual & \ding{51} & - & - & - & - & - & - \\
			MarineInst & 2024 & Image & 2.4M & - & - & Auto-generated & - & - & \ding{51} & - & \ding{51} & - & - \\
			UVLM \cite{xue2025uvlm} & 2025 & Video & 860K & 2.1K & 419 & GPT+Human & - & - & - & - & - & - & \ding{51} \\
			CoralVQA \cite{han2025coralvqa} & 2025 & Image & 12K  &-  &-&GPT+Human& - &- &- &- & -&- & \textbf{\ding{51}}\\
			\textbf{UWBench (Ours)} & \textbf{2025} & \textbf{Image} & \textbf{15K} & \textbf{-} & \textbf{158} & \textbf{GPT+Expert} & \textbf{-} & \textbf{\ding{51}} & \textbf{\ding{51}} & \textbf{-} & \textbf{\ding{51}} & \textbf{\ding{51}} & \textbf{\ding{51}} \\
			\bottomrule
		\end{tabular}
	}
	\label{table1}
\end{table*}

\subsection{Underwater Visual Analysis Tasks}

Research in underwater visual analysis encompasses three primary directions: image quality enhancement \cite{sun2022underwater}, object recognition and localization \cite{elmezain2025advancing}, and environmental modeling \cite{gao2024improved}. 
For image enhancement, numerous methods address the degradation caused by light absorption and scattering in aquatic environments \cite{tian2025adaptive}. 
Physics-based approaches such as Dark Channel Prior and its underwater adaptations model the image formation process to restore color and contrast \cite{du2024physical}. 
Learning-based methods including WaterGAN \cite{li2017watergan}, UGAN \cite{xu2023underwater}, and FUnIE-GAN \cite{islam2020fast} leverage adversarial training for end-to-end enhancement. 
More recent works employ diffusion models \cite{guan2023diffwater} and transformer architectures \cite{peng2023u}, demonstrating improved restoration quality on challenging underwater imagery.

Object recognition in underwater environments has seen substantial progress through specialized detection and classification frameworks \cite{jian2024underwater}. Methods tailored for marine species recognition address challenges including small object scales, camouflage, and inter-species similarity. 
The DeepFish dataset \cite{qin2016deepfish} supports instance segmentation and classification of fish species, while FishNet \cite{khan2023fishnet} provides a large-scale benchmark for fish recognition and functional trait prediction covering over 17,000 species. 
For coral reef analysis, specialized approaches including CoralSCOP \cite{zheng2024coralscop} and semantic segmentation methods enable automated monitoring of reef health and biodiversity. 
Recent vision-language models like MarineInst \cite{zheng2024marineinst} and MarineGPT \cite{zheng2023marinegpt} begin exploring multimodal understanding for marine imagery, incorporating both visual recognition and language description capabilities. While these advances demonstrate significant progress in individual tasks, UWBench uniquely enables integrated evaluation of vision-language capabilities across multiple complementary tasks, providing a holistic framework for assessing comprehensive underwater scene understanding.

\subsection{Multimodal Understanding in Specialized Domains}

The application of vision-language models to specialized domains beyond everyday scenarios has gained increasing attention. In medical imaging, datasets such as MIMIC-CXR \cite{johnson2019mimic} and PadChest \cite{bustos2020padchest} provide radiograph-report pairs enabling the development of models for medical visual question answering and report generation. 
Models like Medunifier \cite{zhang2025medunifier} and BiomedCLIP \cite{zhang2023biomedclip} demonstrate that domain-specific pretraining significantly improves performance on medical tasks compared to generic VLMs. Similarly, in remote sensing, the VRSBench dataset \cite{li2024vrsbench} provides comprehensive annotations including detailed captions, object referring, and visual question answering for aerial imagery. 

These specialized domain applications reveal consistent patterns: domain-specific visual characteristics, specialized vocabulary and knowledge requirements, and limited availability of annotated multimodal data present significant challenges for general-purpose VLMs \cite{guo2025gait, luo2025figvcl, tang2025flag3d++}. 
Successful approaches typically involve constructing large-scale domain-specific datasets, incorporating expert knowledge into annotation processes, and adapting model architectures or training procedures to address domain-specific challenges \cite{huang2025t2i}.
The underwater domain shares these characteristics, exhibiting unique visual degradation patterns, requiring extensive marine biological knowledge, and lacking comprehensive vision-language resources. UWBench follows best practices established in other specialized domains while addressing the unique challenges of underwater environments, providing ecologically informed annotations verified by marine experts and supporting multiple interconnected tasks essential for comprehensive underwater scene understanding. 

\section{UWBench Construction}

\begin{figure*}[t]
	\centering
	\includegraphics[width=1\linewidth]{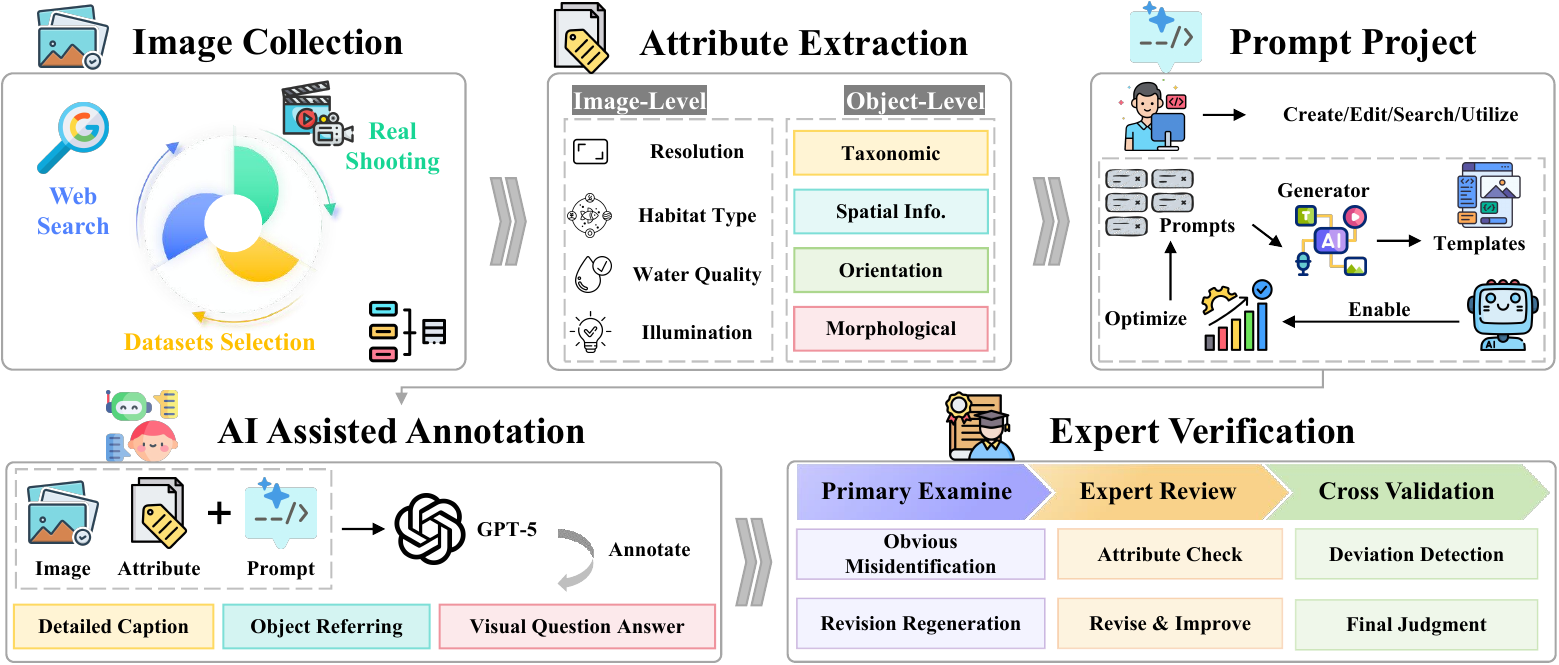} 
	\caption{Overview of UWBench Construction. 
	The pipeline starts with multi-source underwater image acquisition—including web search, public dataset selection, and real-world in-situ shooting—followed by refined attribute extraction that structurally organizes image and object-level features such as resolution, habitat type, water quality, illumination, taxonomy, spatial, and morphological information. 
	Comprehensive prompt engineering then guides GPT-5 to automatically generate detailed captions, distinctive referring expressions, and diverse visual QA pairs. 
	Finally, annotation quality is verified and enhanced through a three-stage process—primary review, expert assessment, and cross-validation—resulting in a scientifically rigorous, ecologically informative, and broadly representative underwater vision-language dataset.
	}
	\label{fig3}
\end{figure*}

\subsection{Overview}

The construction of UWBench addresses three fundamental challenges: capturing underwater visual degradation, incorporating marine biological knowledge, and ensuring annotation quality through expert verification. Our pipeline consists of five stages: data collection, attribute extraction, prompt design, GPT-5 assisted generation, and expert verification. This approach produces 15,003 high-resolution images with 15,281 referring expressions and 124,983 QA pairs, all verified by marine biology experts. Figure \ref{fig3} illustrates the complete annotation pipeline, demonstrating the integration of automated generation and human expertise throughout the construction process.

\subsection{Data Source and Image Collection}

We adopt a multi-source collection strategy drawing from three complementary sources: 
\begin{itemize}
	\item  Web-based collection from Google, Bing, and Flickr targeting marine organisms and underwater habitats, yielding diverse imagery across global locations;
	\item  Samples from existing underwater detection and segmentation datasets providing high-quality instance-level annotations with precise boundaries and taxonomic classifications \cite{hong2025usis16k, islam2020semantic, li2025uwsam, lian2023watermask};
	\item  In-situ images from underwater robotic operations capturing authentic challenges including occlusion and backscatter.
\end{itemize} 
This yields over 35,000 candidate images.

We implement rigorous quality control through systematic filtering by volunteers with marine science backgrounds. The process eliminates severe quality issues, duplicates, watermarked images, and aquarium scenes. Selection criteria emphasize diversity across habitat types (coral reefs, open ocean, kelp forests), water conditions (clear, turbid, low-light), object scales, and scene complexity. Through careful curation, we select 15,003 high-quality images with instance segmentation masks covering 158 object categories.

\subsection{Attribute Extraction}

Attribute extraction transforms segmentation annotations into structured information for language model generation. We extract image-level attributes including resolution, habitat type, water quality indicators, and illumination conditions. Object-level extraction derives taxonomic information (species name, category), spatial information (bounding box, position, relationships), and morphological attributes (size, aspect ratio, visibility status, uniqueness within category). This rich representation provides the foundation for generating diverse vision-language annotations.

\subsection{Prompt Engineering}

We design comprehensive instructions guiding GPT-5 to produce scientifically accurate annotations across three tasks. Our prompts explicitly incorporate marine biological knowledge requirements: proper taxonomic nomenclature, morphological characteristics, behavioral patterns, and ecological context.

\textbf{For image captioning}, prompts instruct GPT-5 to begin with habitat overview including substrate, vegetation, water clarity, and structures, then characterize marine organisms with species identification, morphological features, and positioning. Descriptions use only visually verifiable attributes, avoiding speculation about depth or temperature.

\textbf{For referring expressions}, prompts guide generation of unambiguous descriptions using distinctive visual attributes: morphological features, texture patterns, relative size, coloration, and spatial relationships to visible structures. Each expression must independently identify its target without ordinal references.

\textbf{For visual QA}, prompts elicit diverse question types spanning object identification, existence verification, quantity, shape, size, position, orientation, and scene classification. We generate 3-10 pairs per image with concise answers, prohibiting questions about invisible attributes and requiring exact category names from the taxonomy.

\begin{figure*}[hb]
	\centering
	\includegraphics[width=1\linewidth]{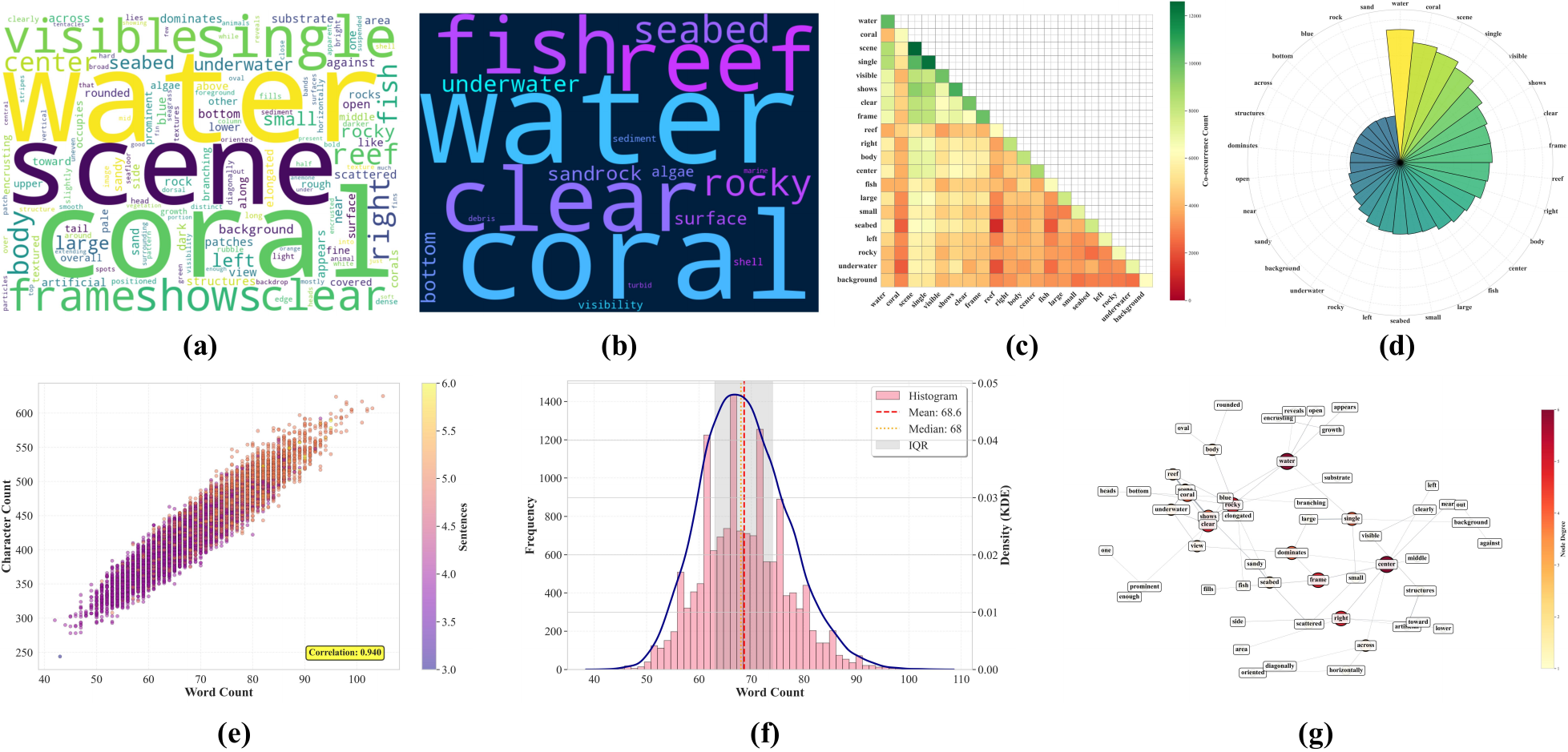} 
	\caption{Statistics of the UWBench caption dataset. 
		(a) Overall Vocabulary Word Cloud.
		(b) Underwater Domain-Specific Terms.
		(c) Top 20 Words Co-occurrence Heatmap.
		(d) Top 30 Words Radial View.
		(e) Word vs Character Count Scatter.
		(f) Word Count Distribution.
		(g) Word Association Network.
	} 
	\label{fig4}
\end{figure*}

\subsection{AI Assisted Annotation Generation}

We employ GPT-5 to generate initial annotations by constructing multimodal prompts combining images, extracted attributes, and task instructions. GPT-5 outputs standardized JSON containing captions, object-referring pairs, and QA pairs.
\begin{itemize}
	\item \textbf{Iterative refinement}: Recursively invoking GPT-5 up to five times to eliminate uncertain language and ambiguous descriptions.
	\item \textbf{Content filtering}: Removing self-answering questions, tautological reasoning, hallucinated information, and questions about invisible properties.
	\item \textbf{Format validation}: Ensuring proper JSON structure, normalized coordinates, and standalone referring expressions.
\end{itemize}
These mechanisms produce high-quality initial annotations for expert verification.

\subsection{Expert Verification}

We implement a rigorous three-stage verification process with marine biology experts:

\begin{itemize}
	\item \textbf{Stage 1: General quality assessment} by trained annotators identifies taxonomic errors, morphological inconsistencies, logical errors, linguistic issues, and guideline violations. Flagged annotations undergo revision or regeneration.
	\item \textbf{Stage 2: Domain expert review} by senior marine biologists validates taxonomic accuracy, ecological plausibility, attribute precision, scientific terminology, and question appropriateness. Experts can modify annotations, add context, or request regeneration.
	\item \textbf{Stage 3: Cross-validation} involves 2,000 images reviewed by multiple experts to assess agreement and identify systematic biases. This reveals error patterns, informs prompt refinement, and establishes quality benchmarks. Senior biologists adjudicate disagreements.
\end{itemize}

The verification process requires approximately 150 seconds per image, totaling over 600 hours. This investment ensures scientifically rigorous annotations with expert-verified taxonomic identifications, ecologically informed descriptions, and diverse question-answer pairs maintaining factual accuracy.

\section{UWBench Statistics}
\subsection{Overview}
UWBench contains 15,003 underwater images with comprehensive human-verified annotations comprising 15,003 detailed captions, 15,281 object referring expressions, and 124,983 visual question-answer pairs. The dataset covers 158 underwater object categories spanning diverse taxonomic groups including marine fishes, shellfish, marine animals, underwater facilities, and debris. All annotations integrate automated GPT-5 generation with rigorous expert verification to ensure scientific accuracy and ecological validity. The detailed statistical results are shown in Table \ref{tab:my_stats}.

\begin{table}[h!]
	\centering
	\caption{Statistics of UWBench.}
	\label{tab:my_stats}
	\renewcommand{\arraystretch}{1.2}
	\resizebox{0.5\textwidth}{!}{
	\begin{tabular}{c|c|c|c|c|c}
		\hline
		\multirow{10}{*}{Caption} & Total Captions & \multirow{10}{*}{\parbox{1.5cm}{\centering Object \\ Referring}} & Total Objects & \multirow{10}{*}{VQA} & Total Pairs \\ \cline{2-2} \cline{4-4} \cline{6-6} 
		& 15,003 & & 15281 & & 124,983 \\ \cline{2-2} \cline{4-4} \cline{6-6} 
		& Avg. Words & & Categories & & Question Types \\ \cline{2-2} \cline{4-4} \cline{6-6} 
		& 68.60 & & 158 & & 301 \\ \cline{2-2} \cline{4-4} \cline{6-6} 
		& Avg. Sentences & & Avg. Objects & & Avg. Q Num \\ \cline{2-2} \cline{4-4} \cline{6-6} 
		& 4.35 & & 1.02 & & 8.33 \\ \cline{2-2} \cline{4-4} \cline{6-6} 
		& Unique Words & & Max Object & & Avg. Q Length \\ \cline{2-2} \cline{4-4} \cline{6-6} 
		& 3560 & & 3 & & 6.91 words \\ \cline{2-2} \cline{4-4} \cline{6-6} 
		& 1st Common Word & & Unique (\%) & & Avg. A Length \\ \cline{2-2} \cline{4-4} \cline{6-6} 
		& water (14,905) & & 99.20\% & & 1.13 words \\
		\hline
	\end{tabular}
}
\end{table}

\begin{figure*}[t]
	\centering
	\includegraphics[width=1\linewidth]{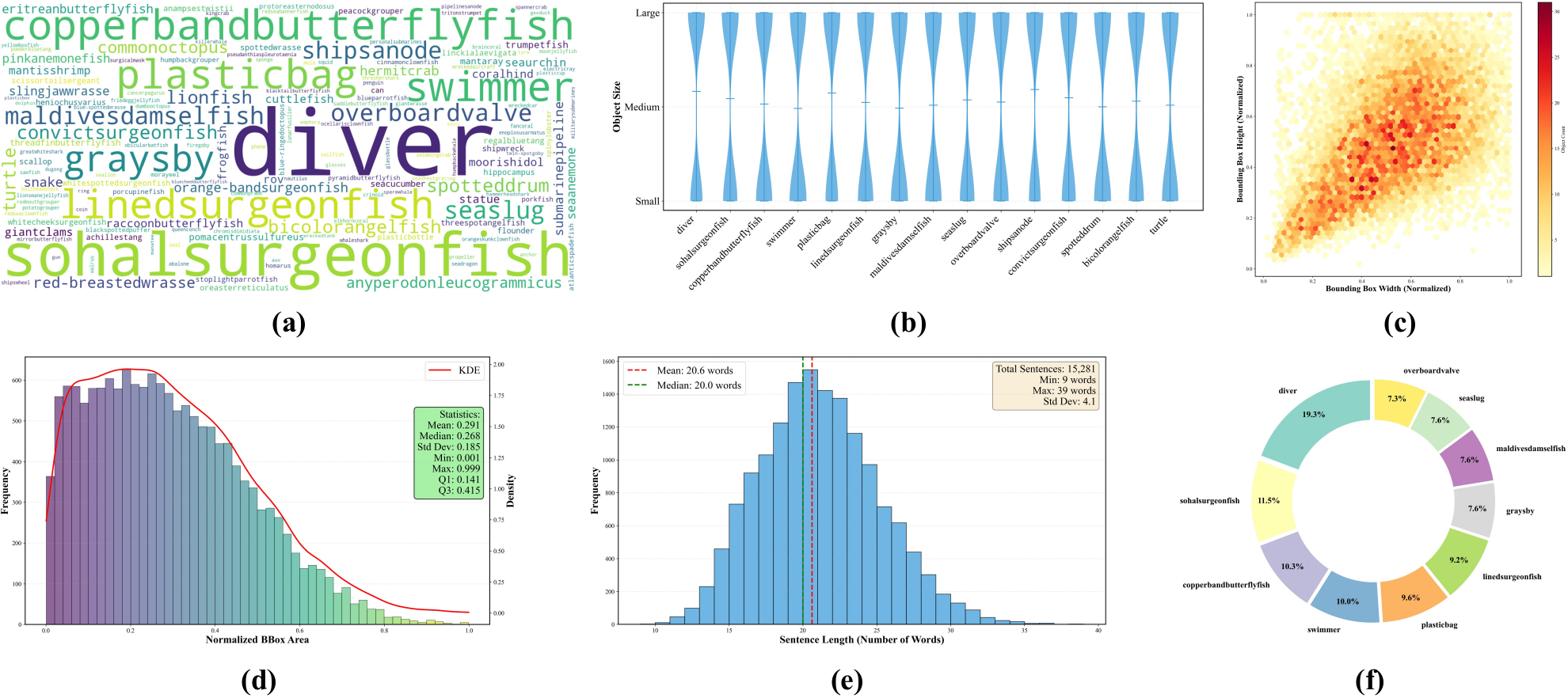} 
	\caption{Statistics of object referring of UWBench. 
		(a) Word Cloud of Categories.
		(b) Object Size by Category (Top 15).
		(c) Density Heatmap for BBox Dimensions.
		(d) BBox Area Distribution.
		(e) Referring Sentence Length Distribution.
		(f) Top 10 Categories Pie Chart (Relative Proportion).
	}
	\label{fig5}
\end{figure*}

\begin{figure*}[ht]
	\centering
	\includegraphics[width=1\linewidth]{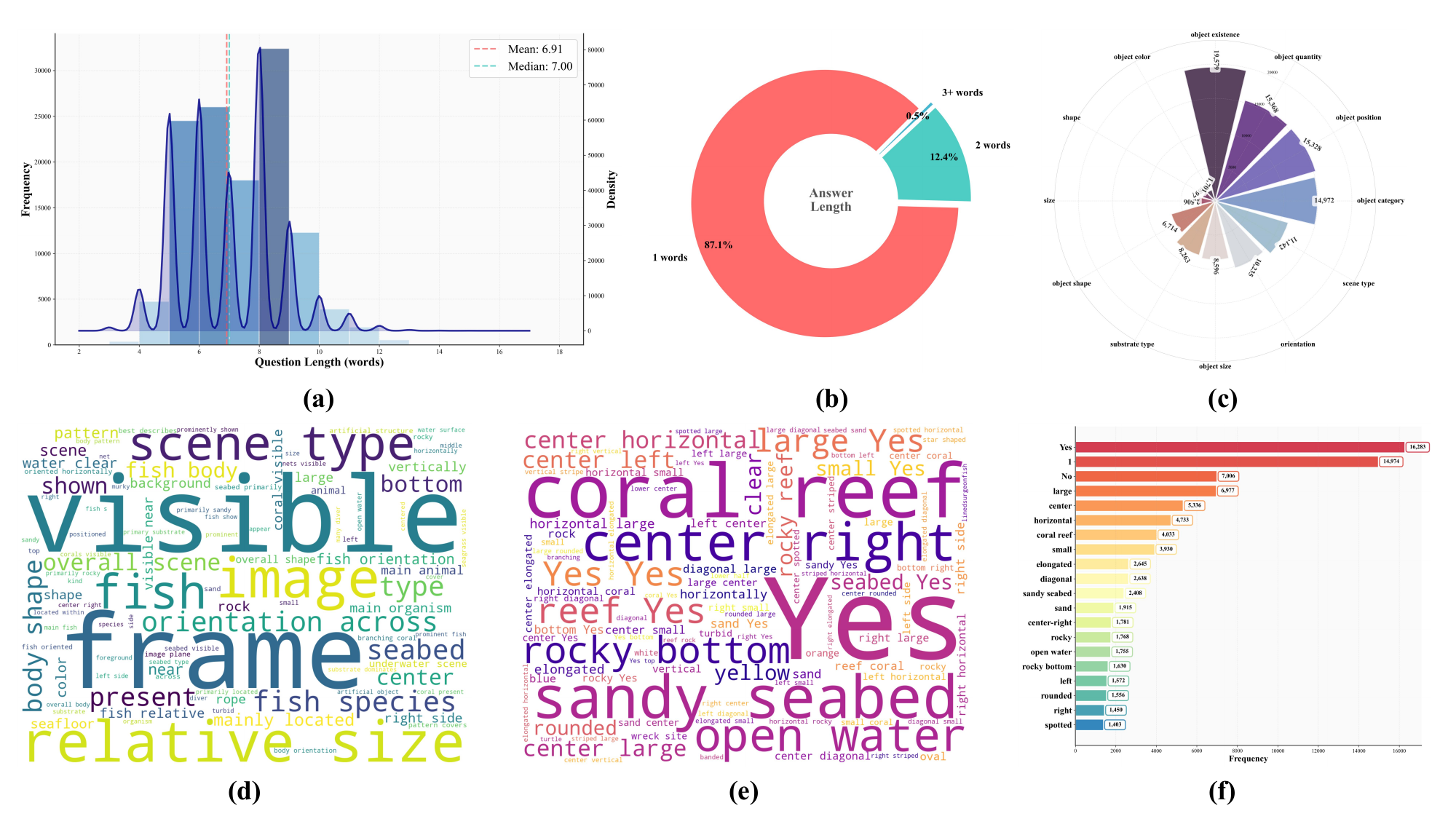} 
	\caption{Statistics of visual question answer pairs in UWBench.
		(a) Question Length Distribution with Density Estimation.
		(b) Answer Length Distribution.
		(c) Question Types - Radial Distribution.
		(d) Question Words Cloud.
		(e) Answer Words Cloud.
		(f) Top 20 Most Frequent Answers.
	}
	\label{fig6}
\end{figure*}

\subsection{Detailed Caption}

UWBench captions provide comprehensive descriptions integrating underwater environmental characteristics with detailed object-specific information. Each caption follows a structured format beginning with scene-level context including substrate type such as sandy seabed, rocky bottom, or coral reef, water quality indicators encompassing clarity, turbidity, and color cast, visible vegetation or algae presence, and artificial structures when present. Following this environmental overview, captions describe prominent marine organisms with attention to species identification, morphological characteristics, relative positioning, and quantity where appropriate. Descriptions emphasize visually verifiable attributes including texture patterns such as spotted or striped markings, shape characteristics, relative size comparisons, and spatial relationships expressed using image-relative terms. Captions explicitly avoid speculation about invisible attributes including exact depth, water temperature, temporal information, or behavioral patterns unless clearly observable.
Statistical analysis reveals captions average 68.60 words across 4.35 sentences, ranging from 30 to 120 words. The dataset vocabulary encompasses 3,560 unique words, with frequent underwater-specific terms including coral, reef, and fish. This reflects the dual focus on environmental context and biological content characteristic of underwater understanding. A summary of these caption statistics is detailed in Figure \ref{fig4}.

\subsection{Object Referring}

UWBench provides referring expressions for 15,281 objects across 158 categories, carefully crafted to enable unambiguous identification without spatial deixis or ordinal references. Each referring sentence independently identifies its target object using distinctive visual attributes rather than relying on positional descriptors or sequential ordering. The expressions emphasize species-specific morphological features, texture patterns including spotted, striped, or encrusted characteristics, relative size compared to other visible organisms, coloration when reliably discernible under underwater lighting conditions, and spatial relationships expressed through adjacency to clearly visible structures such as rocks, coral formations, or artificial objects.
Referring expressions average 20.6 words, ranging from 8 to 40 words. The dataset prioritizes unique objects comprising approximately 60\% of expressions, while remaining expressions address challenging multi-instance scenarios. Category distribution reflects underwater ecosystem composition, which ensures diverse representation across taxonomic groups and object types encountered in underwater observation scenarios. Figure \ref{fig5} provides a summary of UWBench.

\subsection{Visual Question Answering}

UWBench provides 124,983 question-answer pairs covering a variety of reasoning types essential for underwater understanding. 
The questions span object category identification, existence detection, quantity counting, attribute recognition (such as color and shape), spatial relationships, and scene classification (e.g., coral reefs, sandy seabeds). 
Additional questions address substrate and material identification, orientation assessment, and complex reasoning that integrates visual and ecological knowledge, providing a comprehensive evaluation of underwater reasoning abilities.
Answer distribution emphasizes definitive responses with 87.1\% single-word answers, 12.4\% two-word phrases, and 0.5\% three-plus words. Question types extend beyond conventional categories to include underwater-specific reasoning encompassing substrate identification, water quality assessment, organism-substrate relationships, and ecological context inference.
We show the statistics of question-answer pairs in Figure \ref{fig6}.

\section{UWBench Evaluation}
\subsection{Benchmark Overview}

We construct three distinct evaluation tasks to comprehensively assess vision-language capabilities in underwater contexts. UWBench-Cap requires generating comprehensive descriptions for underwater images, capturing environmental characteristics, marine organisms, and their ecological relationships. UWBench-Ref involves identifying and localizing specific underwater objects based on textual descriptions, requiring precise understanding of morphological attributes and spatial relationships. UWBench-VQA aims to answer diverse questions about visual content in underwater scenes, spanning from basic object recognition to complex ecological reasoning.

To facilitate rigorous evaluation, we partition UWBench into non-overlapping training and test splits. The training set comprises 10,454 images with 10,454 captions, 10,654 object referring expressions, and 87,055 question-answer pairs. The test set contains 4,549 images with 4,549 captions, 4,627 object referring expressions, and 37,928 question-answer pairs. This partition ensures that evaluated models demonstrate genuine generalization capability rather than memorization. For this benchmark release, we evaluate state-of-the-art vision-language models exclusively on the test set, providing standardized protocols for fair comparison across different approaches.

We benchmark multiple leading vision-language models including closed-source systems such as GPT-4o, GPT-5, GPT-5-mini, and Gemini 2.5 Flash, alongside open-source models including Qwen2.5-VL series spanning 3B to 72B parameters, InternVL3.5 series ranging from 1B to 241B parameters, Qwen3-VL 30B in both instruction-following and reasoning modes, and GLM-4.1V and GLM-4.5V series. This diverse model selection enables comprehensive analysis of how model architecture, scale, and training methodology affect performance on underwater vision-language understanding tasks.

\begin{table*}[ht]
	\centering
	\caption{Detailed image caption performance on UWBench. Boldface indicates the best performance.}
	\label{caption_result_1}
	\resizebox{1.0\textwidth}{!}{%
		\begin{tabular}{lcccccccc}
			\toprule
			\textbf{Method} & \textbf{BLEU-1} & \textbf{BLEU-2} & \textbf{BLEU-3} & \textbf{BLEU-4} & \textbf{METEOR} & \textbf{ROUGE\_L} &\textbf{ CIDEr} & \textbf{SPICE }\\
			\midrule
			\textbf{GPT-4o} & 40.24 & 22.51 & 12.95 & 7.97 & 22.98 & 26.79 & 33.10 & 28.73 \\
			\textbf{GPT-5} & \textbf{49.17} & \textbf{31.40} & \textbf{21.14} & \textbf{14.90} & \textbf{27.08} & \textbf{34.54} & \textbf{66.10} & \textbf{35.41} \\
			\textbf{GPT-5-mini} & 43.41 & 23.86 & 13.60 & 8.11 & 23.59 & 26.16 & 36.53 & 26.58 \\
			\textbf{Gemini-2.5-Flash} & 33.87 & 16.90 & 8.72 & 4.83 & 23.01 & 22.53 & 10.21 & 24.53 \\
			\midrule
			\textbf{Qwen2.5-VL-3B} & 30.26 & 16.24 & 9.43 & 5.91 & 17.34 & 22.84 & 8.53 & 24.04 \\
			\textbf{Qwen2.5-VL-7B} & 32.60 & 17.31 & 9.83 & 6.06 & 18.17 & 22.85 & 11.87 & 24.72 \\
			\textbf{Qwen2.5-VL-72B} & 40.07 & 22.09 & 12.77 & 7.90 & 21.15 & 25.14 & 31.24 & 27.40 \\
			\midrule
			\textbf{InternVL-3.5-1B} & 31.73 & 17.68 & 10.40 & 6.57 & 17.89 & 22.92 & 11.16 & 24.44 \\
			\textbf{InternVL-3.5-38B} & 36.94 & 20.21 & 11.58 & 7.14 & 20.48 & 24.38 & 26.38 & 26.00 \\
			\textbf{InternVL-3.5-241B} & 35.88 & 19.87 & 11.50 & 7.13 & 20.23 & 24.20 & 23.60 & 26.18 \\
			\midrule
			\textbf{Qwen3-VL-30B-Instruct} & 41.07 & 22.90 & 13.35 & 8.24 & 23.37 & 25.18 & 31.61 & 27.24 \\
			\textbf{Qwen3-VL-30B-Thinking} & 40.82 & 22.64 & 13.28 & 8.19 & 21.74 & 24.86 & 29.22 & 25.79 \\
			\midrule
			\textbf{GLM-4.1V-9B} & 40.28 & 22.21 & 12.87 & 7.87 & 21.15 & 25.02 & 26.40 & 23.74 \\
			\textbf{GLM-4.5V-106B} & 41.96 & 23.41 & 13.67 & 8.41 & 22.45 & 25.46 & 31.48 & 26.68 \\
			\bottomrule
		\end{tabular}%
	}
\end{table*}

\begin{figure*}[ht] 
	\centering
	\begin{subfigure}{.48\textwidth}
		\centering
		\includegraphics[width=\linewidth]{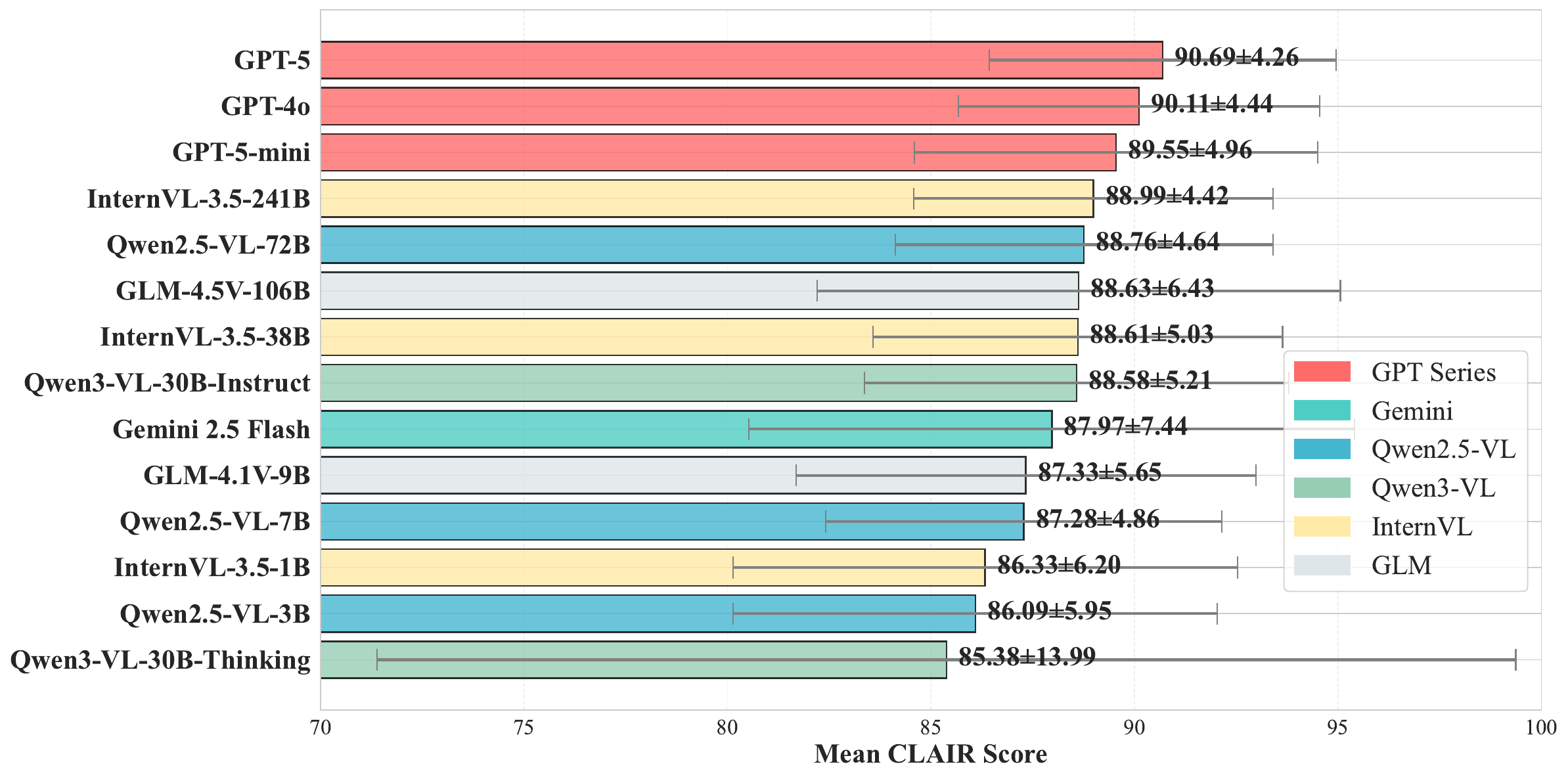}
		\caption{}
		\label{fig:sub1}
	\end{subfigure}%
	\begin{subfigure}{.48\textwidth}
		\centering
		\includegraphics[width=\linewidth]{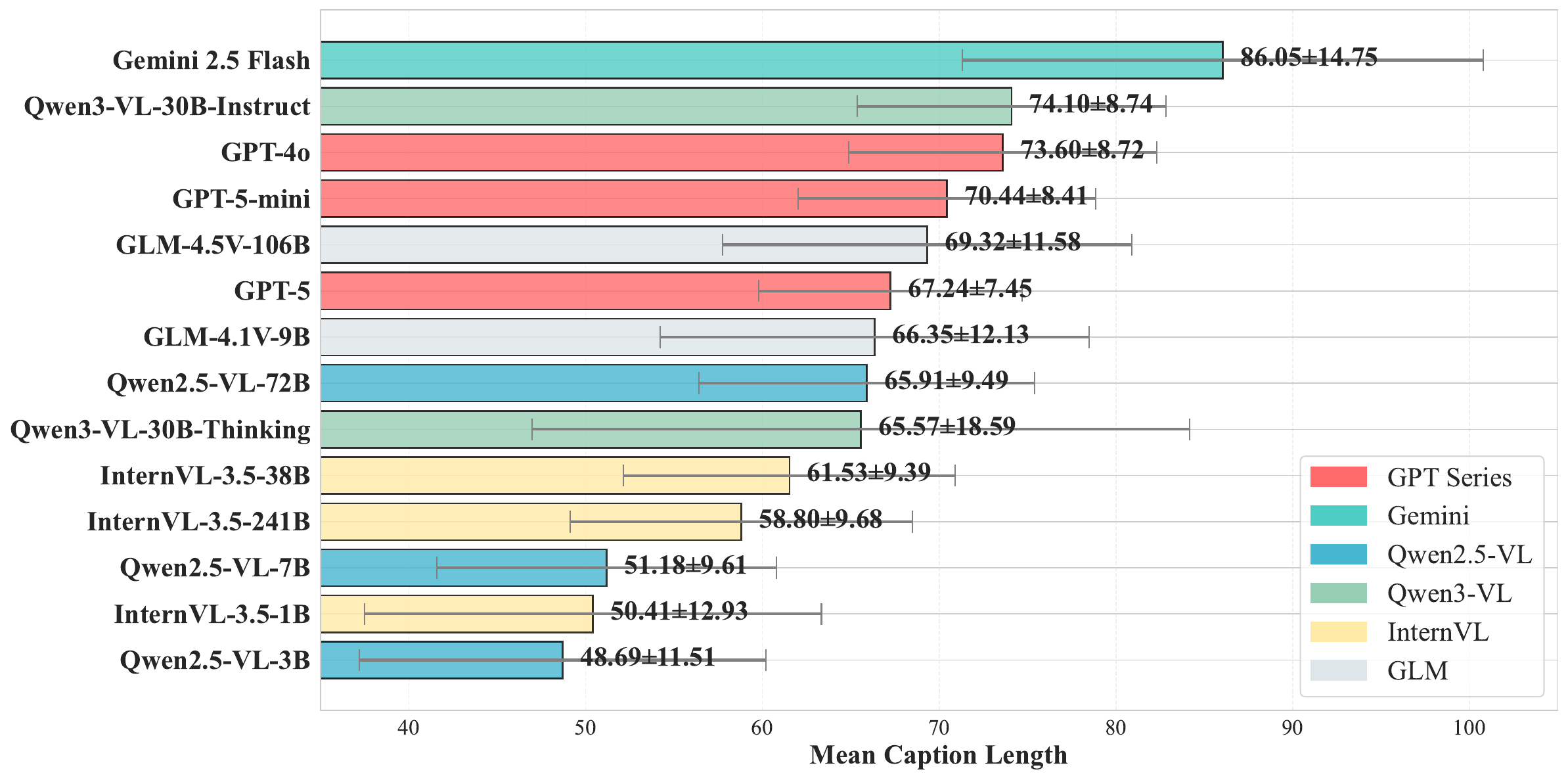}
		\caption{}
		\label{fig:sub2}
	\end{subfigure} \\
	\begin{subfigure}{.48\textwidth}
		\centering
		\includegraphics[width=\linewidth]{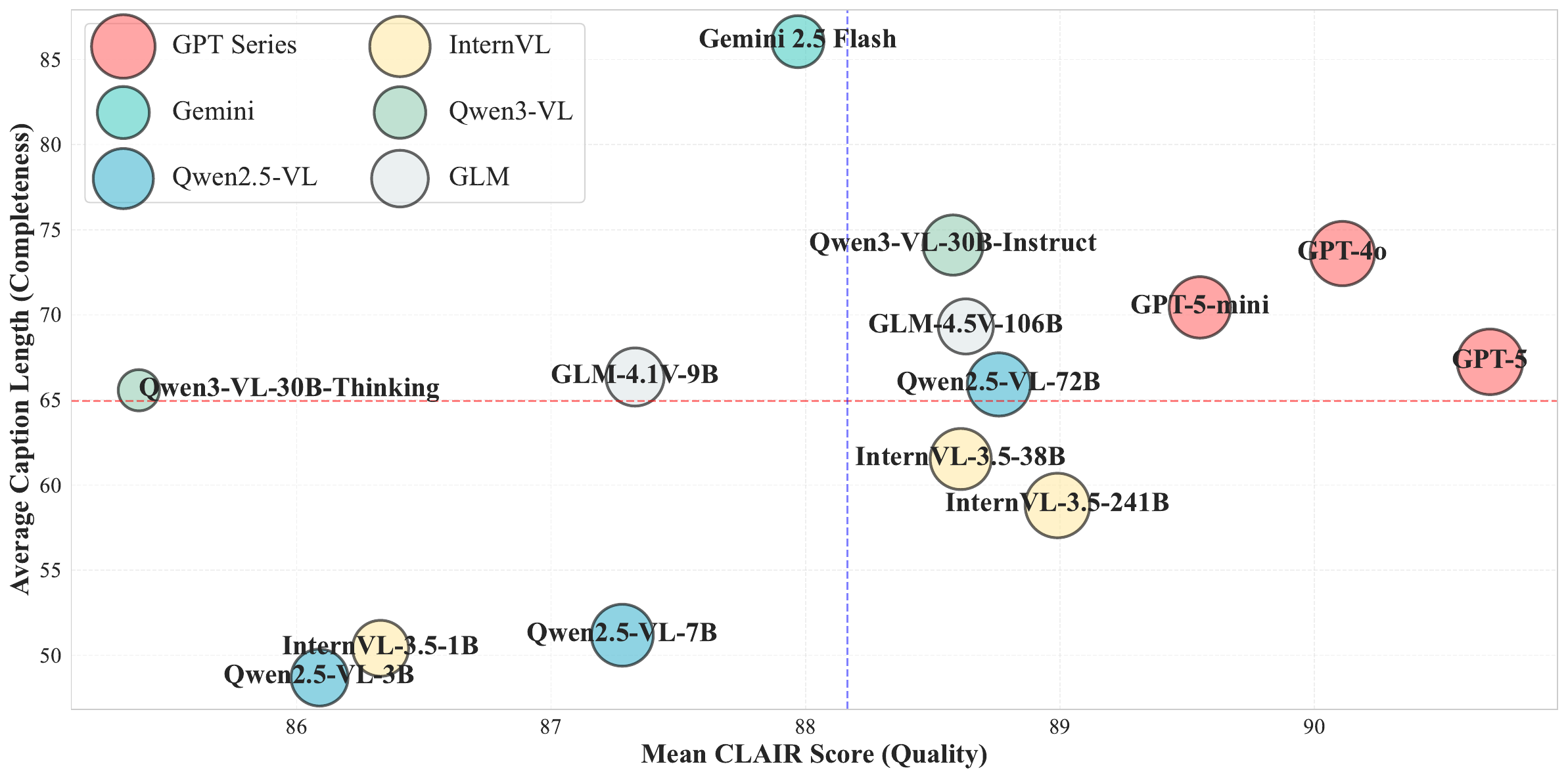}
		\caption{}
		\label{fig:sub3}
	\end{subfigure}%
	\begin{subfigure}{.48\textwidth}
		\centering
		\includegraphics[width=\linewidth]{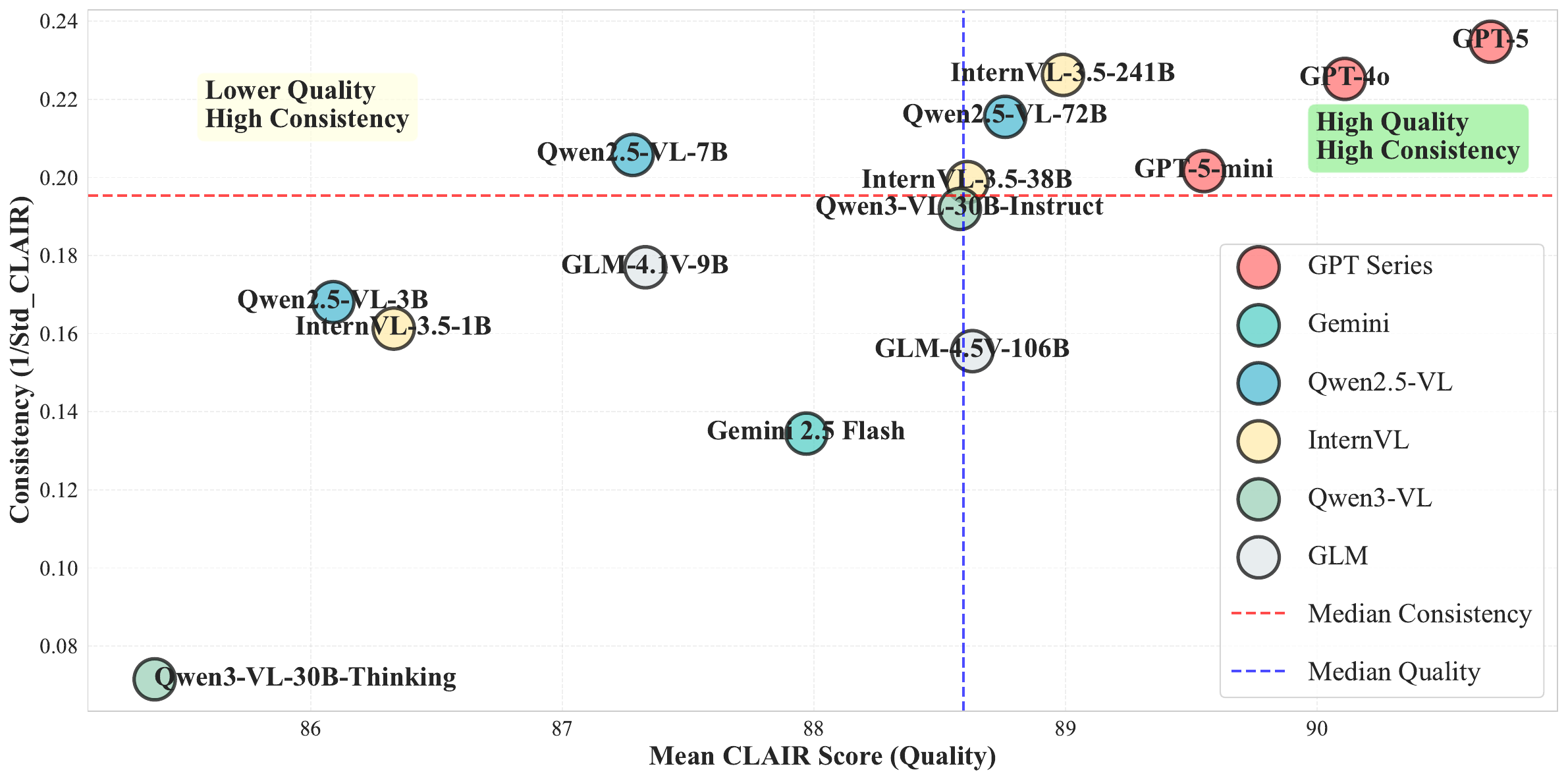}
		\caption{}
		\label{fig:sub4}
	\end{subfigure}
	\caption{(a) Caption Quality: Mean CLAIR Score with Standard Deviation. (b)  Caption Completeness: Mean Caption Length with Standard Deviation. (c) Caption Quality vs Length vs Consistency (Bubble size = Consistency). (d) Quality-Consistency Quadrant Analysis.}
	\label{caption_result_2}
\end{figure*}

\subsection{Detailed Image Caption}

\subsubsection{Settings}

We employ two complementary evaluation approaches to assess caption quality. Traditional automatic metrics provide quantitative measurements of lexical and semantic similarity, while GPT-based evaluation captures higher-level semantic alignment between generated and reference captions.

For automatic evaluation, we utilize established metrics including BLEU measuring n-gram precision with n values of 1 through 4, METEOR assessing semantic similarity through synonym matching and stemming, ROUGE-L computing longest common subsequence, CIDEr evaluating consensus-based image description through TF-IDF weighted n-gram matching, and SPICE measuring semantic propositional content through scene graph matching. These metrics collectively evaluate caption quality from multiple complementary perspectives encompassing lexical precision, semantic similarity, and structural alignment.

We further employ CLAIR score for GPT-based evaluation, which leverages large language models to assess semantic similarity between candidate and reference captions. CLAIR prompts GPT-4o-mini to evaluate on a scale from 0 to 100 how likely a candidate caption describes the same image as the reference caption, considering semantic meaning rather than exact lexical matching. This approach better captures high-level semantic alignment particularly important for detailed underwater captions where diverse valid phrasings exist for describing complex scenes. We report mean CLAIR score, standard deviation, and median across the test set. Additionally, we compute average caption length and standard deviation to assess description comprehensiveness.

\begin{figure*}[hbpt]
	\centering
	\includegraphics[width=1\linewidth]{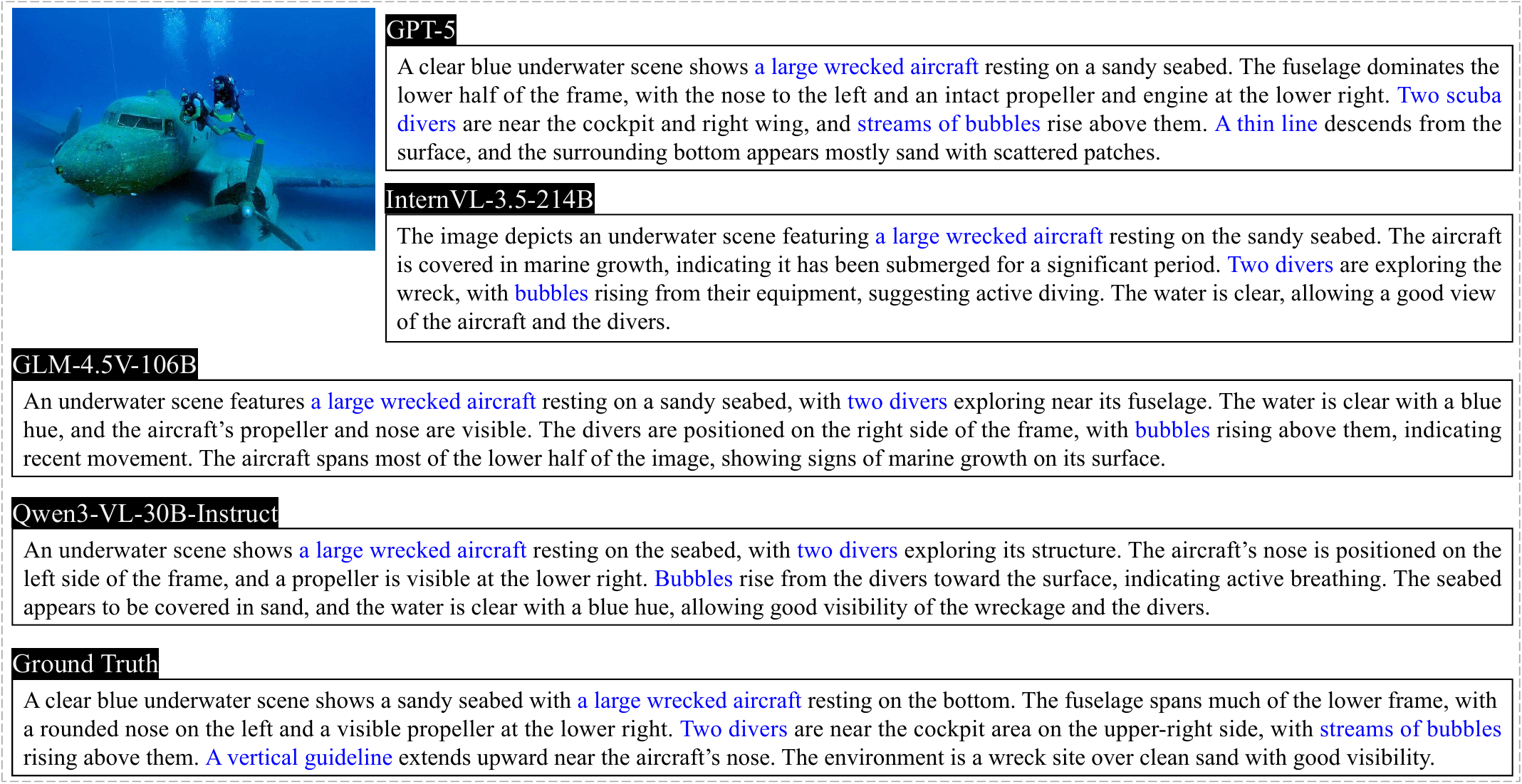} 
	\caption{Selected examples of detailed image caption results. We highlight pivotal information in \textcolor{blue}{blue}.
	}
	\label{fig8}
\end{figure*}

\subsubsection{Results}

Experimental results (Table \ref{caption_result_1}) demonstrate substantial variation in model performance across underwater image captioning. GPT-5 achieves the highest automatic metric scores with BLEU-4 of 14.90, METEOR of 27.08, and CIDEr of 66.10, substantially outperforming other approaches. GPT-4o and GPT-5-mini follow with competitive performance, while Gemini 2.5 Flash exhibits lower automatic metric scores despite generating longer captions averaging 86.05 words. Among open-source models, GLM-4.5V and Qwen3-VL-30B-Instruct demonstrate strong performance approaching closed-source systems, while smaller models including Qwen2.5-VL-3B and InternVL-3.5-1B show significant performance gaps.

As figure \ref{caption_result_2} shows, GPT-based CLAIR evaluation reveals consistently high semantic alignment across most models. GPT-5 achieves the highest CLAIR score of 90.69, followed closely by GPT-4o at 90.11 and GPT-5-mini at 89.55. Open-source models demonstrate competitive CLAIR performance with GLM-4.5V, InternVL-3.5-241B, and InternVL-3.5-38B all exceeding 88.5, indicating strong semantic understanding despite gaps in automatic metrics. Notably, Qwen3-VL-30B-Thinking exhibits lower CLAIR score of 85.38 with high standard deviation of 13.99, suggesting less consistent caption quality in reasoning mode.
Analysis of caption length reveals interesting patterns. Closed-source models generally produce captions closer to reference length of 68.6 words, with GPT-4o generating 73.60 words and Qwen3-VL-30B-Instruct producing 74.10 words. Gemini 2.5 Flash generates substantially longer captions averaging 86.05 words, while smaller open-source models produce more concise descriptions ranging from 48.69 to 61.53 words. This suggests that caption comprehensiveness positively correlates with model scale and training methodology, with larger models better capturing the detailed environmental and biological information characteristic of underwater scenes.

\begin{figure*}[hbpt]
	\centering
	\includegraphics[width=0.95\linewidth]{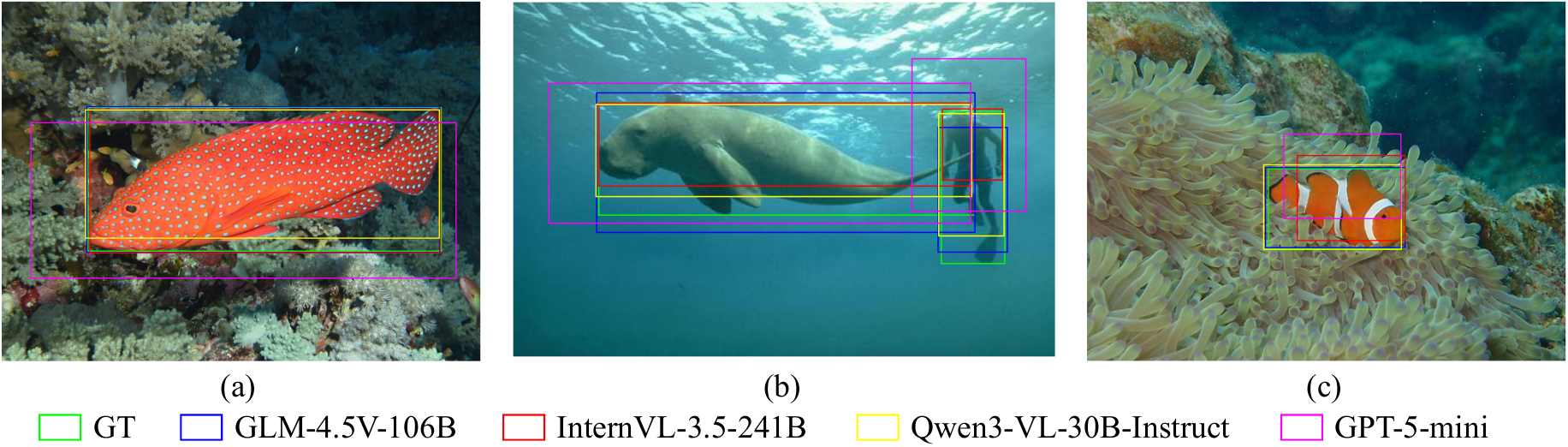} 
	\caption{Selected examples of referring object. 
		(a) The large elongated coralhind covered in small pale spots extends horizontally across the middle of the image above the corals.
		(b) The dugong stretching horizontally across the middle of the image just below the bright water surface. \& The lone diver on the right side of the frame, positioned to the right of the dugong.
		(c) The small ocellarisclownfish partly hidden among pale tubular anemone tentacles in the center-right of the image.
	}
	\label{fig9}
\end{figure*}

\subsection{Object Referring}

\begin{table}[htbp]
	\centering
	\caption{Object referring performance on UWBench. Boldface indicates the best performance.},
	\label{referring_result}
	\resizebox{0.5\textwidth}{!}{%
		\begin{tabular}{lcccc}
			\toprule
			\textbf{Method} & \textbf{Acc@IoU\_0.5} & \textbf{Acc@IoU\_0.7} & \textbf{mIoU} & \textbf{Cum\_IoU} \\
			\midrule
			\textbf{GPT-4o} & 28.29 & 6.78 & 36.78 & 43.64 \\
			\textbf{GPT-5} & 62.81 & 22.78 & 54.22 & 60.37 \\
			\textbf{GPT-5-mini} & 70.37 & 24.83 & 56.20 & 62.37 \\
			\textbf{Gemini-2.5-Flash} & 60.21 & 17.53 & 52.97 & 57.70 \\
			\midrule
			\textbf{Qwen2.5-VL-3B} & 54.55 & 15.37 & 47.34 & 51.60 \\
			\textbf{Qwen2.5-VL-7B} & 37.15 & 11.76 & 32.77 & 33.69 \\
			\textbf{Qwen2.5-VL-72B} & 90.90 & 74.71 & 77.90 & \textbf{81.50} \\
			\midrule
			\textbf{InternVL-3.5-1B} & / & / & / & / \\
			\textbf{InternVL-3.5-38B} & 55.91 & 22.36 & 51.38 & 32.53 \\
			\textbf{InternVL-3.5-241B} & 84.76 & 64.14 & 71.99 & 76.98 \\
			\midrule
			\textbf{Qwen3-VL-30B-Instruct} & \textbf{94.40} & \textbf{80.38} & \textbf{80.18} & 51.67 \\
			\textbf{Qwen3-VL-30B-Thinking} & 56.52 & 30.04 & 50.78 & 56.00 \\
			\midrule
			\textbf{GLM-4.1V-9B} & 85.95 & 68.92 & 75.27 & 48.07 \\
			\textbf{GLM-4.5V-106B} & 89.21 & 75.38 & 79.26 & 80.51 \\
			\bottomrule
		\end{tabular}%
	}
\end{table}

\subsubsection{Settings}
Visual grounding evaluation assesses model capability to localize underwater objects based on textual referring expressions. We focus on bounding box prediction using normalized coordinates where models must generate location specifications in the format of four corner coordinates representing top-left and bottom-right positions.

We employ Intersection over Union based accuracy metrics to evaluate localization performance. Accuracy at threshold tau measures the proportion of predictions where IoU between predicted and ground truth bounding boxes exceeds tau. We report Acc@0.5 and Acc@0.7 representing accuracy at IoU thresholds of 0.5 and 0.7 respectively, providing evaluation at both moderate and strict localization criteria. Additionally, we compute mean IoU across all predictions to assess average localization precision, and cumulative IoU measuring the average maximum IoU achievable, indicating the upper bound of model localization capability.

\subsubsection{Results}

Table \ref{referring_result} reveal substantial performance variation across models in underwater object grounding. Qwen3-VL-30B-Instruct achieves the highest Acc@0.5 of 94.40\% and Acc@0.7 of 80.38\%, demonstrating exceptional localization precision. Qwen2.5-VL-72B and GLM-4.5V follow with strong performance exceeding 89\% at IoU 0.5 threshold, while InternVL-3.5-241B and GLM-4.1V achieve competitive accuracy above 84\%. Among closed-source models, GPT-5-mini surprisingly outperforms GPT-5 with 70.37\% accuracy at threshold 0.5, while GPT-4o exhibits substantially lower performance at 28.29\%, suggesting that grounding capability does not necessarily scale with general vision-language performance.

Mean IoU analysis reveals consistent patterns with accuracy metrics. Top-performing models including Qwen3-VL-30B-Instruct, GLM-4.5V, and Qwen2.5-VL-72B achieve mIoU exceeding 77\%, indicating precise bounding box predictions. Mid-tier models range from 50\% to 60\% mIoU, while GPT-4o demonstrates significantly lower precision at 36.78\%. Cumulative IoU measurements show similar trends, with best models exceeding 80\% and representing strong upper bound performance potential.

Analysis indicates that open-source models generally outperform closed-source systems in visual grounding tasks on underwater imagery. This contrasts with caption generation results and suggests that grounding benefits particularly from domain-specific fine-tuning on object localization tasks. The substantial gap between GPT-5 and GPT-5-mini further indicates that general-purpose training may not optimally transfer to precise spatial reasoning in underwater contexts. Smaller models including Qwen2.5-VL-7B and InternVL-3.5-38B demonstrate moderate performance, highlighting the importance of model scale for complex spatial understanding in challenging underwater conditions.

\begin{figure}[ht]
	\centering
	\includegraphics[width=1\linewidth]{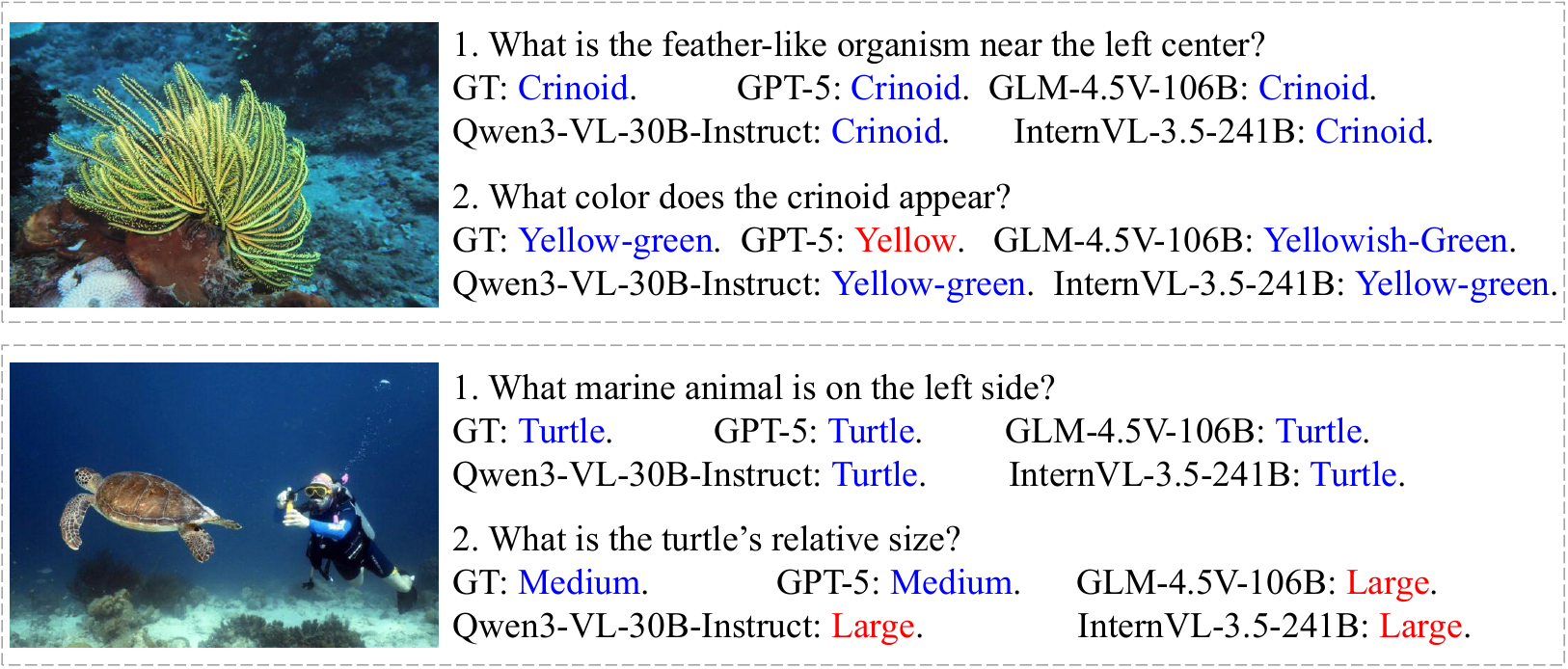} 
	\caption{Selected examples of VQA results. Correct answers are shown in \textcolor{blue}{blue} and incorrect answers are shown in \textcolor{red}{red}.
	}
	\label{fig10}
\end{figure}

\subsection{Visual Question Answering}

\begin{table*}[htbp]
	\centering
	\caption{Visual question answering performance on UWBench. Boldface indicates the best performance.}
	\label{vqa_result}
	\resizebox{1.0\textwidth}{!}{%
		\begin{tabular}{lccccccccccc}
			\toprule
			\textbf{Method} 
			& \shortstack[c]{\textbf{Object} \\ \textbf{Identification}} 
			& \shortstack[c]{\textbf{Quantity} \\ \& \textbf{Existence}} 
			& \shortstack[c]{\textbf{Position} \& \\ \textbf{Spatial Relations}} 
			& \shortstack[c]{\textbf{Shape, Size} \\ \& \textbf{Form}} 
			& \shortstack[c]{\textbf{Color, Texture} \\ \& \textbf{Pattern}} 
			& \shortstack[c]{\textbf{Object Attributes} \\ \& \textbf{Features}} 
			& \shortstack[c]{\textbf{Scene} \& \\ \textbf{Environment}} 
			& \shortstack[c]{\textbf{Substrate} \\ \& \textbf{Materials}} 
			& \shortstack[c]{\textbf{Water Quality} \\ \& \textbf{Visibility}} 
			& \shortstack[c]{\textbf{Reasoning} \\ \& \textbf{Comparison}} 
			& \textbf{All} \\
			\midrule
			\textbf{VQAs} & 4558 & 10618 & 8063 & 6102 & 1679 & 160 & 3540 & 2604 & 363 & 241 & 37928 \\
			\midrule
			\textbf{GPT-4o} & \textbf{99.98} & 99.58 & 89.94 & 83.26 & 95.11 & 95.25 & 66.92 & 90.63 & 89.26 & 95.43 & 90.95 \\
			\textbf{GPT-5} & 99.96 & \textbf{99.61} & 90.28 & 85.04 & \textbf{97.32} & \textbf{98.75} & \textbf{85.48} & \textbf{93.78} & \textbf{95.04} & \textbf{97.51} & \textbf{93.44} \\
			\textbf{GPT-5-mini} & 97.28 & 96.96 & 85.02 & 79.24 & 93.45 & 93.75 & 82.09 & 90.74 & 86.50 & 93.77 & 89.50 \\
			\textbf{Gemini-2.5-Flash} & 14.52 & 1.47 & 61.55 & 71.42 & 45.26 & 24.37 & 47.74 & 47.58 & 40.77 & 26.14 & 37.12 \\
			\midrule
			\textbf{Qwen2.5-VL-3B} & 38.22 & 34.46 & 30.03 & 35.41 & 39.73 & 39.37 & 30.08 & 33.91 & 41.87 & 44.40 & 34.06 \\
			\textbf{Qwen2.5-VL-7B} & 13.84 & 49.91 & 55.66 & 64.42 & 38.95 & 21.87 & 44.89 & 42.24 & 35.54 & 23.24 & 39.06 \\
			\textbf{Qwen2.5-VL-72B} & 99.91 & 9.54 & 88.45 & \textbf{93.22} & 93.03 & 98.75 & 76.04 & 90.09 & 86.22 & 96.26 & 91.04 \\
			\midrule
			\textbf{InternVL-3.5-1B} & 14.70 & 1.77 & 58.75 & 70.11 & 42.29 & 23.75 & 47.06 & 43.47 & 38.01 & 26.97 & 35.91 \\
			\textbf{InternVL-3.5-38B} & 99.38 & 99.12 & 89.62 & 82.59 & 92.97 & 96.87 & 69.43 & 88.63 & 87.05 & 95.85 & 90.56 \\
			\textbf{InternVL-3.5-241B} & 99.96 & 99.43 & 90.09 & 84.66 & 93.92 & 96.87 & 68.67 & 91.09 & 88.43 & 96.26 & 91.31 \\
			\midrule
			\textbf{Qwen3-VL-30B-Instruct} & 99.96 & 99.16 & \textbf{90.44} & 85.17 & 95.12 & 96.87 & 71.89 & 89.82 & 88.71 & 96.27 & 91.66 \\
			\textbf{Qwen3-VL-30B-Thinking} & 13.23 & 39.55 & 61.07 & 71.83 & 43.84 & 25.62 & 48.31 & 45.16 & 39.12 & 26.14 & 41.39 \\
			\midrule
			\textbf{GLM-4.1V-9B} & 15.40 & 2.33 & 61.27 & 71.98 & 45.02 & 28.12 & 47.32 & 47.17 & 39.94 & 26.97 & 37.43 \\
			\textbf{GLM-4.5V-106B} & 21.87 & 8.73 & 62.48 & 72.52 & 48.78 & 27.50 & 50.62 & 49.85 & 40.22 & 29.05 & 41.01 \\
			\bottomrule
		\end{tabular}%
	}
\end{table*}

\subsubsection{Settings}

VQA evaluation assesses model capability to answer diverse questions about underwater scenes. We employ GPT-based semantic matching to determine answer correctness, comparing predicted and ground truth answers through semantic similarity rather than exact string matching. This approach accommodates the semantic equivalence of synonymous terms common in underwater domain such as seabed and seafloor, or turbid and murky.

We utilize GPT-4o-mini to evaluate answer similarity through a carefully designed prompt that provides the question, ground truth answer, and predicted answer, instructing the model to determine whether predicted answer matches ground truth considering semantic meaning. The model returns binary judgment of 1 for match or 0 for no match. To improve efficiency, we implement rule-based matching for straightforward cases including exact matches and simple containment relationships, reserving GPT evaluation for ambiguous cases requiring semantic reasoning.
To enable fine-grained analysis, we categorize questions into ten major types reflecting diverse reasoning capabilities required for comprehensive underwater understanding. Detailed classification results are shown in the Appendix.

\subsubsection{Results}

Results in Table \ref{vqa_result} reveal substantial variation across question categories and model capabilities. GPT-5 achieves the highest overall accuracy of 93.44\%, demonstrating strong performance across all categories with particular strength in Object Identification at 99.96\%, Quantity and Existence at 99.61\%, and Color, Texture, and Pattern at 97.32\%. GPT-4o follows closely with 90.95\% overall accuracy, while GPT-5-mini achieves competitive 89.50\%. Among open-source models, performance varies dramatically. InternVL-3.5-241B achieves the highest accuracy at 91.31\%, closely followed by Qwen3-VL-30B-Instruct at 91.66\% and Qwen2.5-VL-72B at 91.04\%, approaching closed-source performance. However, substantial performance gaps emerge across question categories. All top-performing models exceed 99\% on Object Identification and Quantity tasks but show significant challenges on Scene and Environment questions where accuracy drops to 68\% to 86\%, indicating difficulty in holistic scene reasoning.

Smaller open-source models demonstrate markedly lower performance. GLM-4.5V achieves 41.01\% overall accuracy, while GLM-4.1V and Qwen3-VL-30B-Thinking reach approximately 37\% to 41\%. These models exhibit severe difficulty on Quantity and Existence questions, with accuracy dropping below 10\% in some cases, while achieving moderate performance exceeding 60\% on Position and Shape categories. This suggests smaller models struggle particularly with numerical reasoning and abstract existence verification but retain reasonable capability for concrete spatial and morphological understanding.
Category-specific analysis reveals consistent patterns. Object Identification proves relatively easier with most models exceeding 95\% except smaller variants. Quantity and Existence shows the largest performance gap, with top models approaching perfect accuracy while smaller models fail dramatically, likely due to the precise numerical reasoning required for counting underwater organisms. Position and Spatial Relations demonstrates moderate difficulty with top models achieving approximately 90\% while smaller models reach 60\%. Shape, Size, and Form follows similar patterns. Water Quality and Visibility shows intermediate difficulty around 86\% to 95\% for strong models, indicating that environmental assessment benefits from both visual perception and domain knowledge.

\subsection{Visualization}

We have visualized the three UWBench tasks, demonstrating the typical performance of mainstream models in underwater scene vision-language understanding:

For image caption (Figure \ref{fig8}), the generated results from various models are generally highly consistent with the manually annotated reference descriptions. They all accurately capture key scene elements, such as the wreckage of the sunken aircraft, the sandy bottom, the diver's bubbles, and the clear blue water. They also describe some spatial structure (such as the nose orientation, propeller position, and diver distribution). However, there are subtle differences in performance between different models: For example, InternVL focuses on describing biological traces attached to the aircraft, while GLM and Qwen3-VL-30B-Instruct provide more complete depictions of details and visible structures, with clearer spatial relationships. GPT-5's descriptions are the most comprehensive and closely match the human-annotated descriptions.

For object referring (Figure \ref{fig9}), the models showed good consistency in selecting and localizing objects (e.g., aircraft, propellers, divers, etc.). Visualization results show a high degree of overlap in localization or highlighting across both the ground truth and the models, demonstrating that mainstream models possess accurate underwater object understanding and spatial localization capabilities. Fine-grained differences primarily manifest in accuracy under complex structures or occlusion.

For vqa task (Figure \ref{fig10}), the models achieved high agreement with the ground truth answers for most questions, particularly for specific organisms (e.g., the feathery creature is a crinoid, the animal on the left is a sea turtle), demonstrating strong species recognition. They also performed well on color and relative size questions. For example, with the exception of GPT-5, the large models' judgment of the "yellow-green" color scheme was highly consistent with the ground truth answer. While some models exhibited slight bias on individual questions, overall, all models provided reasonable and scientific answers. This demonstrates that the large models have achieved a high level of fine-grained attribute recognition and reasoning capabilities in underwater scenes.

\section{Conclusion}

We introduce UWBench, the first large-scale comprehensive benchmark specifically designed for underwater vision-language understanding. The benchmark comprises 15,003 high-resolution underwater images with 15,003 human-verified detailed captions, 15,281 object referring expressions, and 124,983 visual question-answer pairs spanning 158 underwater object categories. Through a rigorous semi-automatic construction pipeline combining GPT-5 assisted generation with multi-stage expert verification, we ensure scientific accuracy and ecological validity across all annotations. UWBench facilitates comprehensive evaluation across three interconnected tasks including detailed image captioning, visual grounding, and visual question answering, providing standardized protocols for systematic assessment of vision-language models in challenging aquatic environments. Our extensive evaluation of state-of-the-art vision-language models reveals that underwater image understanding remains highly challenging even for the most advanced systems. These findings highlight the critical need for specialized approaches tailored to underwater contexts, motivating continued research in this important yet underexplored domain.

\textbf{Limitations and Future Work}: Current vision-language models exhibit common weaknesses including difficulty with precise numerical reasoning, limited understanding of complex ecological relationships, and challenges in integrating domain-specific knowledge with visual perception. Our ongoing efforts include fine-tuning specialized vision-language models on UWBench training data to develop systems with enhanced underwater understanding capabilities.

\bibliographystyle{IEEEtran}
\bibliography{reference}

\end{document}